\documentclass{article}

\usepackage{iclr2019_conference,times}

\usepackage[utf8]{inputenc} 
\usepackage[T1]{fontenc}    
\usepackage{hyperref}       
\usepackage{url}            
\usepackage{booktabs}       
\usepackage{amsmath}
\usepackage{amsfonts}       
\usepackage{nicefrac}       
\usepackage{microtype}      
\usepackage[pdftex]{graphicx}
\usepackage{placeins}
\usepackage{color}

\newcommand{\argmin}{\operatornamewithlimits{argmin}}

\usepackage{amssymb}
\usepackage{mathtools}
\usepackage{amsmath}
\usepackage{hyperref}

\usepackage{xcolor}
\definecolor{darkgreen}{rgb}{0,0.6,0}
\definecolor{darkred}{rgb}{0.7,0.0,0}
\definecolor{darkblue}{rgb}{0,0.0,0.6}

\title{Adversarial Reprogramming of \\ Neural Networks}

\author{Gamaleldin F. Elsayed\thanks{Work done as a member of the Google AI Residency program (g.co/airesidency).} \\
  Google Brain\\
  \texttt{gamaleldin.elsayed@gmail.com}
  \And
Ian Goodfellow \\
Google Brain \\
\texttt{goodfellow@google.com}
\And
Jascha Sohl-Dickstein\\
Google Brain\\
\texttt{jaschasd@google.com}
}

\iclrfinalcopy 
\begin{document}

\maketitle

\begin{abstract}
Deep neural networks are susceptible to \emph{adversarial} attacks. 
In computer vision, well-crafted perturbations to images can cause
neural networks to make mistakes such as confusing a cat with a computer.
Previous adversarial attacks have been designed to degrade performance
of models or cause machine learning models to produce specific outputs chosen
ahead of time by the attacker.
We introduce attacks that instead {\em reprogram} the target model
to perform a task chosen by the attacker---without the attacker needing
to specify or compute the desired output for each test-time input.
This attack finds a single adversarial perturbation, that can be added to all test-time inputs to
a machine learning model in order to cause the model to perform a task chosen by
the adversary---even if the model was not
trained to do this task.
These perturbations can thus be considered a program for the new task. 
We demonstrate adversarial reprogramming on six ImageNet classification models,
repurposing these models to perform a counting task, as well as classification
tasks: classification of MNIST and CIFAR-10 examples presented as inputs
to the ImageNet model.

\end{abstract}

\section{Introduction}

The study of adversarial examples is often motivated in terms of the danger posed by an attacker whose goal is to cause model prediction errors with a small change to the model's input. 
Such an attacker could make a self-driving car react to a phantom stop sign \citep{evtimov2017robust} by means of a sticker (a small $L_{0}$ perturbation), or cause an insurance company's damage model to overestimate the claim value from the resulting accident by subtly doctoring photos of the damage (a small $L_\infty$ perturbation). 
With this context, 
various methods have been proposed both to construct \citep{szegedy2013intriguing, papernot2015limitations, papernot2017practical, papernot2016transferability, brown2017adversarial, liu2016delving} and defend against
 \citep{goodfellow2014explaining,kurakin2016mlatscale,madry2017towards,ensemble_training, kolter2017provable, kannan2018adversarial} this style of adversarial attack. 
Thus far, the majority of adversarial attacks have consisted of {\em untargeted}
 attacks that aim to degrade the performance of a model without necessarily
 requiring it to produce a specific output, or {\em targeted} attacks
 in which the attacker designs an adversarial perturbation to produce
 a specific output for that input.
 For example, an attack against a classifier might target a specific desired output class for
 each input image, or an attack against a reinforcement learning agent might induce
 that agent to enter a specific state \citep{lin2017tactics}.

In practice, there is no constraint that adversarial attacks should adhere to this framework. Thus, it is crucial to proactively anticipate other unexplored adversarial goals in order to make machine learning systems more secure. 
In this work, we consider a novel and more challenging adversarial goal:
reprogramming the model to perform a task chosen by the attacker,
without the attacker needing to compute the specific desired output.
Consider a model trained to perform some {\em original task}:
for inputs $x$ it produces outputs $f(x)$.
Consider an adversary who wishes to perform an {\em adversarial task}:
for inputs $\tilde{x}$ (not necessarily in the same domain as $x$)
the adversary wishes to compute a function $g(\tilde{x})$.
We show that an adversary can accomplish this by learning 
{\em adversarial reprogramming functions}
$h_f(\cdot ; \theta)$ and $h_g(\cdot ; \theta)$ that map 
between the two tasks.
Here, $h_f$ converts 
inputs from the domain of $\tilde{x}$ 
into the domain of $x$ 
(i.e.,
$h_f(\tilde{x}; \theta)$ is a valid input to the function $f$), 
while $h_g$ maps output of $f(h(\tilde{x}; \theta))$ back to outputs of $g(\tilde{x})$. 
The parameters $\theta$ of the adversarial program are then adjusted to achieve
$h_g\left(
	f\left(
		h_f\left(\tilde{x}\right)
	\right)
\right) = g\left(
	\tilde{x}
\right)$.
%
%

In our work, for simplicity, 
we define $\tilde{x}$ to be a small image,
$g$ a function that processes small images,
$x$ a large image,
and $f$ a function that processes large images.
Our function $h_f$ then just consists of drawing $x$ in the center
of the large image and $\theta$ in the borders (though we explore other schemes as well), and $h_g$ is simply a hard coded mapping between output class labels. 
However, the idea is more general; $h_f$ ($h_g$) could be any consistent
transformation that converts between the input (output) formats for the two tasks
and causes the model to perform the adversarial task.

We refer to the class of attacks where a model is repurposed to
perform a new task as {\em adversarial reprogramming}.
We refer to $\theta$
as an {\em adversarial program}.
In contrast to most previous adversarial work, the magnitude of this perturbation need not be constrained for adversarial reprogramming to work. Though, we note that it is still possible to construct reprogramming attacks that are imperceptible.
Potential consequences of adversarial reprogramming include theft of computational resources from public facing services, repurposing of AI-driven assistants into spies or spam bots, and abusing machine learning services for tasks violating the ethical principles of system providers. 
Risks stemming from this type of attack are discussed in more detail in Section 
\ref{sec beyond images}.

It may seem unlikely that an additive offset to a neural network's input would be sufficient on its own to repurpose the network to a new task. 
However, this flexibility stemming 
only from changes to a network's inputs is consistent with results on the expressive power of deep neural networks. 
For instance, in \citet{raghu2016expressive} it is shown that, depending on network hyperparameters, the number of unique output patterns achievable by moving along a one-dimensional trajectory in input space increases exponentially with network depth. 
%

In this paper, we present the first instances of adversarial reprogramming. 
In Section \ref{sec related}, we discuss related work. 
In Section \ref{sec method}, we present a training procedure for crafting adversarial programs, which cause a neural network to perform a new task. 
In Section \ref{sec results}, we experimentally demonstrate adversarial programs that target several convolutional neural networks designed to classify ImageNet data. 
These adversarial programs alter the network function from ImageNet classification to: counting squares in an image, classifying MNIST digits, and classifying CIFAR-10 images. 
Next, we examine the susceptibility of trained and untrained networks to adversarial reprogramming. We then demonstrate the possibility of reprograming adversarial tasks with adversarial data that has no resemblance to original data, demonstrating that results from transfer learning do not fully explain adversarial reprogramming. Further, we demonstrate the possibility of concealing adversarial programs and data.
Finally, we end in Sections \ref{sec discuss} and \ref{sec conclusion} by discussing and summarizing our results.


\begin{figure}
\centering
\includegraphics[width=\columnwidth]{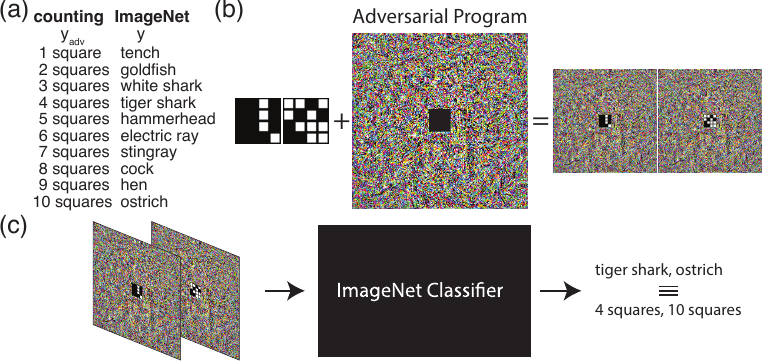}
\caption{{\bf Illustration of adversarial reprogramming.}
(a) Mapping of ImageNet labels to adversarial task labels (squares count in an image). (b) Two examples of images from the adversarial task (left) are embedded at the center of an adversarial program (middle), yielding adversarial images (right). The adversarial program shown repurposes an Inception V3 network to count squares in images.
(c) Illustration of inference with adversarial images. The network when presented with adversarial images will predict ImageNet labels that map to the adversarial task labels.}
\label{fig: counting}
\end{figure}

\section{Background and Related Work}\label{sec related}
 
\subsection{Adversarial examples}

One definition of adversarial examples is that they are
``inputs to machine learning models that an attacker has intentionally designed to cause the model to make a mistake''
\citep{goodfellow2017}.
They are often formed by starting with a naturally occuring image
and using a gradient-based optimizer to search for a nearby
image that causes a mistake
\citep{Biggio13,szegedy2013intriguing,7958570}.
These attacks can be either {\em untargeted} (the adversary succeeds
when causing any mistake at all) or {\em targeted} (the adversary
succeeds when causing the model to predict a specific incorrect class).
Adversarial attacks have been also proposed for other domains like malware detection \citep{grosse17}, generative models \citep{kos2017adversarial}, network policies for reinforcement learning tasks \citep{huang2017adversarial}, and network interpretations \citep{ghorbani2017interpretation}. 
In these domains, the attack remains either untargeted (generally degrading the performance)
or targeted (producing a specific output).
We extend this line of work by developing reprogramming methods that aim to
produce specific {\em functionality} rather than a specific hardcoded output.

Several authors have observed that the same modification can be
applied to many different inputs in order to form adversarial
examples \citep{goodfellow2014explaining,moosavi2017universal}.
For example,  \citet{brown2017adversarial} designed an
``adversarial patch'' that can switch the prediction of many models
to one specific class (e.g. toaster) when it is placed physically
in their field of view.
We continue this line of work by finding a single adversarial
program that can be presented with many input images to cause
the model to process each image according to the adversarial
program.





\subsection{Parasitic Computing and Weird Machines}

Parasitic computing involves 
forcing a target system to solve a complex computational task it wasn't originally designed to perform, 
by taking advantage of peculiarities in network communication protocols \citep{barabasi2001parasitic,peresini2013network}. 
Weird machines, on the other hand, are a class of computational exploits where carefully crafted inputs can be used to run arbitrary code on a targeted computer \citep{bratus2011exploit}. 
Adversarial reprogramming can be seen as a form of parasitic computing, though without the focus on leveraging the communication protocol itself to perform the computation. 
Similarly, adversarial reprogramming can be seen as an example of neural networks behaving like weird machines, 
though adversarial reprogramming functions only within the neural network paradigm -- we do not gain 
access to the host computer.

\subsection{Transfer Learning}

Transfer learning \citep{raina2007self, mesnil2011unsupervised} and adversarial reprogramming share the goal of repurposing networks to perform a new task. 
Transfer learning methods use the knowledge obtained from one task as a base to learn how to perform another. Neural networks possess properties that can be useful for many tasks \citep{yosinski2014transferable}. For example, neural networks when trained on images develop features that resemble Gabor filters in early layers even if they are trained with different datasets or different training objectives such as supervised image classification \citep{krizhevsky2012imagenet}, unsupervised density learning \citep{lee2009convolutional}, or unsupervised learning of sparse representations \citep{le2011ica}. Empirical work has demonstrated that it is possible to take a convolutional neural network trained to perform one task, and simply train a linear SVM classifier to make the network work for other tasks \citep{razavian2014cnn, donahue2014decaf}. 
However, transfer learning is very different from the adversarial reprogramming task in that it allows model parameters to be changed for the new task. In typical adversarial settings, an attacker is unable to alter the model, and instead must achieve their goals solely through manipulation of the input. Further, one may wish to adversarially reprogram across tasks with very different datasets. This makes the task of adversarial reprogramming much more challenging than transfer learning. 

\section{Methods}\label{sec method}

In this work, we consider an adversary with access to the parameters of a neural network that is performing a specific task. The objective of the adversary is to reprogram the model to perform a new task by crafting an adversarial program to be included within the network input. Here, the network was originally designed to perform ImageNet classification, but the methods discussed here can be directly extended to other settings.

Our adversarial program is formulated as an additive contribution to network input. 
Note that unlike most adversarial perturbations, the adversarial program is not specific to a single image. 
The same adversarial program will be applied to all images.
We define the adversarial program as:
\begin{align}
P = \tanh\left(W \odot M\right)
\end{align}
where $W \in \mathbb{R}^{n \times n \times 3}$ is the adversarial program parameters to be learned, $n$ is the ImageNet image width, and $M$ is a masking matrix that is 0 for image locations that corresponds to the adversarial data for the new task, otherwise 1. 
Note that the mask $M$ is not required -- we mask out the central region of the adversarial program purely to improve visualization of the action of the adversarial program. 
Also, note that we use $\tanh\left(\cdot\right)$ to bound the adversarial perturbation to be in $(-1, 1)$ -- the same range as the (rescaled) ImageNet images the target networks are trained to classify. 

Let, $\tilde{x} \in \mathbb{R}^{k \times k \times 3}$ be a sample from the dataset to which we wish to apply the adversarial task, where $k < n$. $\tilde{X} \in \mathbb{R}^{n \times n \times 3}$ is the equivalent ImageNet size image with $\tilde{x}$ placed in the proper area, defined by the mask $M$.
The corresponding adversarial image is then:
\begin{align}
X_{adv} = h_f\left(\tilde{x}; W\right) = 
\tilde{X} + P
\label{eqn: adv prog}
\end{align}

Let $P(y | X)$ be the probability that an ImageNet classifier gives to ImageNet label $y \in \{1, \hdots, 1000\}$, given an input image $X$. We define a hard-coded mapping function $h_g(y_{adv})$ that maps a label from an adversarial task $y_{adv}$ to a set of ImageNet labels. For example, if an adversarial task has 10 different classes ($y_{adv} \in \{1, \hdots, 10\}$), $h_g\left(\cdot\right)$ may be defined to assign the first 10 classes of ImageNet, any other 10 classes, or multiple ImageNet classes to the adversarial labels. Our adversarial goal is thus to maximize the probability $P(h_g(y_{adv}) | X_{adv})$. We set up our optimization problem as
\begin{align}
\label{eq adv obj}
\hat{W} = 
\argmin_{W} \left( 
	- \log P(h_g(y_{adv}) | X_{adv}) + \lambda ||W||_F^2
	\right)
,
\end{align}
where $\lambda$ is the coefficient for a weight norm penalty, to reduce overfitting.
We optimize this loss with Adam while exponentially decaying the learning rate. Hyperparameters 
are given in Appendix \ref{sec: supp tables}.
Note that after the optimization the adversarial program has a minimal computation cost from the adversary's side as it only requires computing $X_{adv}$ (Equation \ref{eqn: adv prog}), and mapping the resulting ImageNet label to the correct class. 
In other words, during inference the adversary needs only to store the program and add it to the data, thus leaving the majority of computation to the target network. 

One interesting property of adversarial reprogramming is that it must
exploit nonlinear behavior of the target model.
This is in contrast to traditional adversarial examples, where attack
algorithms based on linear approximations of deep neural networks are
sufficient to cause high error rate \citep{goodfellow2014explaining}.
Consider a linear model that receives an input $\tilde{x}$ and a program
$\theta$ concatenated into a single vector: $x = [\tilde{x}, \theta]^{\top}$.
Suppose that the weights of the linear model are partitioned into two
sets, $v = [v_{\tilde{x}}, v_\theta]^{\top}$.
The output of the model is $v^{\top} x = v_{\tilde{x}}^{\top} \tilde{x} + v_\theta^{\top} \theta$.
The adversarial program $\theta$ adapts the effective biases
$v_\theta^{\top} \theta$ but cannot adapt the weights applied to the input $\tilde{x}$.
The adversarial program $\theta$ can thus bias the model toward consistently outputting one class or
the other but cannot change the way the input is processed.
For adversarial reprogramming to work, the model must include nonlinear interactions of $\tilde{x}$ and $\theta$.
A nonlinear deep network satisfies this requirement.

\begin{figure}
\centering
\includegraphics[width=\columnwidth]{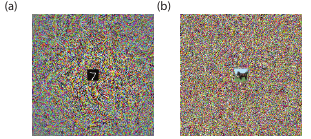}
\caption{{\bf Examples of adversarial programs.}
Adversarial program which cause Inception V3 ImageNet model to function as (a) MNIST classifier. (b) CIFAR-10 classifier}
\label{fig: ex}
\end{figure}

\section{Results}\label{sec results}
To demonstrate the feasibility of adversarial reprogramming, 
we conducted experiments on six architectures trained on ImageNet. 
In each case, we reprogrammed the network to perform three different adversarial tasks: counting squares, MNIST classification, and CIFAR-10 classification. 
The weights of all trained models were obtained from \citet{tfslim}, and top-1 ImageNet precisions are shown in Table \ref{table: ImageNet precision}. 
We additionally examined whether adversarial training conferred resistance to adversarial reprogramming, and compared the susceptibility of trained networks to random networks. Further, we investigated the possibility of reprogramming the networks when the adversarial data has no resemblance to the original data. Finally, we demonstrated the possibility of concealing the adversarial program and the adversarial data.

\subsection{Counting squares}

To illustrate the adversarial reprogramming procedure, we start with a simple adversarial task. That is counting the number of squares in an image. 
We generated images ($\tilde x$) of size $36 \times 36 \times 3$ that include $9 \times 9$ white squares with black frames. Each square could appear in 16 different position in the image, and the number of squares ranged from $1$ to $10$. The squares were placed randomly on gridpoints (Figure \ref{fig: counting}b left). 
We embedded these images in an adversarial program (Figure \ref{fig: counting}b middle). 
The resulting images ($X_{adv}$) are of size $299 \times 299 \times 3$ with the $36 \times 36 \times 3$ images of the squares at the center (Figure \ref{fig: counting}b right). Thus, the adversarial program is simply a frame around the counting task images. 
We trained one adversarial program per ImageNet model, such that the first 10 ImageNet labels represent the number of squares in each image (Figure \ref{fig: counting}c). 
Note that the labels we used from ImageNet have no relation to the labels of the new adversarial task. For example, a `White Shark' has nothing to do with counting 3 squares in an image, and an `Ostrich' does not at all resemble 10 squares. 
We then evaluated the accuracy in the task by sampling 100,000 images and comparing the network prediction to the number of squares in the image.

Despite the dissimilarity of ImageNet labels and adversarial labels, and that the adversarial program is equivalent simply to a first layer bias, the adversarial program masters this counting task for all networks (Table \ref{table: reprogramming results}). 
These results demonstrate the vulnerability of neural networks to reprogramming on this simple task using only additive contributions to the input.


\subsection{MNIST classification}
\label{sec mnist}

\begin{table}
  \caption{{\bf Neural networks adversarially reprogrammed to perform a variety of tasks.}
  Table gives accuracy of reprogrammed networks to perform a counting task, MNIST classification task, and CIFAR-10 classification task, and Shuffled MNIST pixels classification task.}
  \label{table: reprogramming results}
  \centering
  \begin{tabular}{llllllll}
    \toprule
       Model &  \multicolumn{6}{c}{Pretrained on ImageNet}   & \multicolumn{1}{c}{Untrained}       \\
        \cmidrule(r){2-7}
        \cmidrule(r){8-8}

   &Counting &  \multicolumn{2}{c}{MNIST}     &       \multicolumn{2}{c}{CIFAR-10}     & \multicolumn{1}{c}{Shuffled MNIST}  & \multicolumn{1}{c}{MNIST}       \\
        \cmidrule(r){3-4}
        \cmidrule(r){5-6}
        \cmidrule(r){7-7}
        \cmidrule(r){8-8}

           &  & train &  test & train & test  &test  &test  \\
    \midrule
    Incep. V3                 &  $0.9993$  &   $0.9781$    & $0.9753$    & $0.7311$  & $0.6911$  & $0.9709$ & $0.4539$  \\
    Incep. V4                 & $0.9999$   &    $0.9638$  & $0.9646$   &  $0.6948$   & $0.6683$& $0.9715$ & $0.1861$  \\
    Incep. Res. V2    & $0.9994$   &   $0.9773$  & $0.9744$    &  $0.6985$ & $0.6719$ & $0.9683$ & $0.1135$  \\
    Res. V2 152             &  $0.9763$  &    $0.9478$ &$0.9534$   &  $0.6410$ & $0.6210$ & $0.9691$ & $0.1032$ \\
    Res. V2 101             &  $0.9843$  &  $0.9650$  & $0.9664$  & $0.6435$  & $0.6301$ & $0.9678$ & $0.1756$ \\
    Res. V2 50             &  $0.9966$  &  $0.9506$  & $0.9496$  & $0.6$  & $0.5858$ & $0.9717$ &  $0.9325$ \\
    Incep. V3 adv.          &     &   $0.9761$    & $0.9752$    &   &  & & \\
    \bottomrule
  \end{tabular}
\end{table}


In this section, we demonstrate adversarial reprogramming on somewhat more complex task of classifying MNIST digits. We measure {\em test} and train accuracy, so it is impossible for the adversarial program to have simply memorized all training examples.
Similar to the counting task, we embedded MNIST digits of size $28 \times 28 \times 3$ inside a frame representing the adversarial program, we assigned the first 10 ImageNet labels to the MNIST digits, and trained an adversarial program for each ImageNet model. 
Figure \ref{fig: ex}a shows example of the adversarial program for Inception V3 being applied. 

Our results show that ImageNet networks can be successfully reprogramed to function as an MNIST classifier by presenting an additive adversarial program. 
The adversarial program additionally generalized well from the training to test set, suggesting that the reprogramming does not function purely by memorizing train examples, and is not brittle to small changes in the input. 

\subsection{CIFAR-10 classification}

Here we implement a more challenging adversarial task. That is, crafting adversarial programs to repurpose ImageNet models to instead classify CIFAR-10 images. 
An example of the resulting adversarial images are given in Figure \ref{fig: ex}b.
Our results show that our adversarial program was able to increase the accuracy on CIFAR-10 from chance to a moderate accuracy (Table \ref{table: reprogramming results}). This accuracy is near what is expected from typical fully connected networks \citep{lin2015far} but with minimal computation cost from the adversary side at inference time. 
One observation is that although adversarial programs trained to classify CIFAR-10 are different from those that classify MNIST or perform the counting task, the programs show some visual similarities, e.g. ResNet architecture adversarial programs seem to possess some low spatial frequency texture (Figure \ref{fig: programs}a).


\subsection{Investigation of the effect of the trained model details and original data}

One important question is what is the degree to which susceptibility to adversarial reprogramming depends on the details of the model being attacked. 
To address this question, we examined attack success on an Inception V3 model that was trained on ImageNet data using adversarial training \citep{ensemble_training}. Adversarial training augments data with adversarial examples during training, and is one of the most common methods for guarding against adversarial examples. 
As in Section \ref{sec mnist}, we adversarially reprogrammed this network to classify MNIST digits.
Our results (Table \ref{table: reprogramming results}) indicate that the model trained with adversarial training is still vulnerable to reprogramming, with only a slight reduction in attack success. 
This finding shows that standard approaches to adversarial defense has little efficacy against adversarial reprogramming.
\begin{figure}
\centering
\includegraphics[width=\columnwidth]{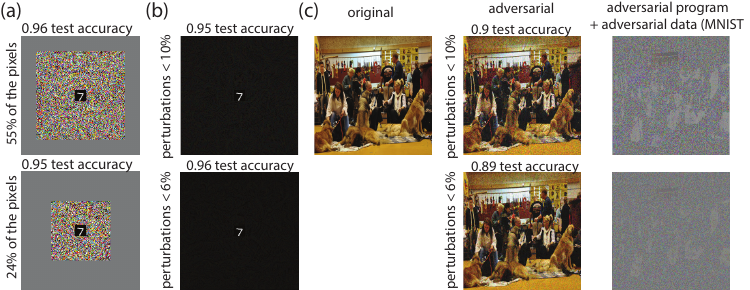}
\caption{{\bf Adversarial programs may be limited in size or concealed.}
In all panels, an Inception V3 model pretrained on ImageNet is reprogrammed to classify MNIST digits. Example images
(a) with adversarial programs of different sizes.
(b) with adversarial programs of different perturbation scales.
(c) Here adversarial data + program (right) are hidden inside a normal image from ImageNet (left), yielding an adversarial image (center) that is able to reprogram the network to function as an MNIST classifier. The pixels of the adversarial data are shuffled to conceal its structure.
}
\label{fig: concealing programs}
\end{figure}
This is likely explained by the differences between adversarial reprogramming and standard adversarial attacks. First, that the goal is to repurpose the network rather than cause it to make a specific mistake, second that the magnitude of adversarial programs can be large, while traditional adversarial attacks are of a 
small perturbations magnitude, and third adversarial defense methods may be specific to original data and may not generalize to data from the adversarial task.




To further explore dependence on the details of the model, we performed adversarial reprogramming attacks on models with random weights.
We used the same experiment set up and MNIST reprogramming task as in Section \ref{sec mnist} -- we simply used the ImageNet models with randomly initialized rather than trained weights.
The MNIST classification task was easy for networks pretrained on ImageNet (Table \ref{table: reprogramming results}). However, for random networks, training was very challenging and generally converged to a much lower accuracy (only ResNet V2 50 could train to a similar accuracy as trained ImageNet models; see Table \ref{table: reprogramming results}). Moreover, the appearance of the adversarial programs was qualitatively distinct from the adversarial programs obtained with networks pretrained on ImageNet (see Figure \ref{fig: programs}b). This finding demonstrates that the original task the neural networks perform is important for adversarial reprogramming.   
 This result may seem surprising, as random networks have rich structure adversarial programs might be expected to take advantage of. 
 For example, theoretical results have shown that wide neural networks become identical to Gaussian processes, where training specific weights in intermediate layers is not necessary to perform tasks \citep{matthews2018gaussian,lee2017deep}. Other work has demonstrated that it is possible to use random networks as generative models for images \citep{ustyuzhaninov2016texture, he2016powerful}, further supporting their potential richness. One explanation may be that randomly initialized networks perform poorly for simple reasons, such as poor scaling of network weights at initialization, whereas the trained weights are better conditioned. 

One explanation of adversarial reprogramming that is motivated by transfer learning  \citep{yosinski2014transferable} is that the network may be relying on some similarities between original and adversarial data. To address this hypothesis, we randomized the pixels on MNIST digits such that any resemblance between the adversarial data (MNIST) and images in the original data (ImageNet) is removed (see Figure \ref{fig: shuffled}). We then attempted to reprogram pretrained ImageNet networks to classify the shuffled MNIST digits. Despite shuffled MNIST not sharing any spatial structure with images, we managed to reprogram the ImageNet network for this task (Table \ref{table: reprogramming results}) with almost equal accuracy to standard MNIST (in some cases shuffled MNIST even achieved higher accuracy). These results thus suggest that transferring knowledge between the original and adversarial data does not explain the susceptibility to adversarial reprogramming. Even more interestingly, these results suggest the possibility of reprogramming across tasks with unrelated datasets and across domains.

\subsection{Concealing adversarial programs}
In our previous experiment, there were no constraints on the size (number of program pixels) and scale (magnitude of perturbations) of the program and adversarial data. Here, we demonstrate the possibility of limiting the visibility of the adversarial perturbations by limiting the program size, scale, or even concealing the whole adversarial task. In these experiments, we used Inception V3 model pretrained to classify ImageNet. In our first experiment, we adversarially reprogrammed the network to classify MNIST digits while limiting the size of the program (see Figure \ref{fig: concealing programs}a). Our results show that adversarial reprogramming is still successful, yet with lower accuracy, even if we used a very small adversarial program. In our next experiment, we made the adversarial program nearly imperceptible by limiting the $L_{\inf}$ norm of the adversarial perturbation to a small percentage of the pixel values. Our results show that adversarial reprogramming is still successful (see Figure \ref{fig: concealing programs}b) even with nearly imperceptible programs. Further, we tested the possibility of concealing the whole adversarial task by hiding both the adversarial data and program within a normal image from ImageNet. To do that, we shuffled the pixels of the adversarial data (here MNIST), so that the adversarial data structure is hidden. Then, we limited the scale of both the adversarial program and data to a small fraction of the possible pixel values. We added the resulting image to a random image from ImageNet. Formally, we extended our reprogramming methods as follows:
\begin{align}
P_X &= \alpha \tanh\left(\text{shuffle}_{ix}(\tilde X) + (W \odot \text{shuffle}_{ix}(M)) \right) \\
X_{adv} &= \text{clip}\left(X_{ImageNet} +P_X, ~~[0, 1] \right)
,
\end{align}
where $\tilde X$, $M$ and $W$ are as described in Section \ref{sec method}, $P_X$ is the adversarial data combined with the adversarial program, $ix$ is the shuffling sequence (same for $M$ and $\forall X$), $\alpha$ is a scalar used to limit the perturbation scale, and $X_{ImageNet}$ is an image chosen randomly from ImageNet, which is the same for all MNIST examples. We then optimized the adversarial program for the network to classify MNIST digits (see Equation \ref{eq adv obj}). The resulting adversarial images are very similar to normal images from ImageNet (see Figure \ref{fig: concealing programs}c), yet the network is successfully reprogrammed to classify MNIST digits, though with lower accuracy (see Figure \ref{fig: concealing programs}c). This result demonstrate the possibility of hiding the adversarial task. Here, we used a simple shuffling technique and picked an image from ImageNet to hide the adversarial task, but one could go further and use more complex schemes for hiding the adversarial task and optimize the choice of the image from ImageNet, which may make adversarial reprogramming even harder to detect.

\section{Discussion}\label{sec discuss}

\subsection{Flexibility of trained neural networks}
%
%
%
%
%

We found that trained neural networks were more susceptible to adversarial reprogramming than random networks. Further, we found that reprogramming is still successful even when data structure is very different from the structure of the data in the original task.
This demonstrates a large flexibility of repurposing trained weights for a new task. 
Our results suggest that dynamical reuse of neural circuits should 
be practical in modern artificial neural networks. 
This holds the promise of enabling machine learning systems which are easier to repurpose, more flexible, and more efficient due to shared compute. 
Indeed, recent work in machine learning has focused on building large dynamically connected networks with reusable components \citep{shazeer2017outrageously}.

It is unclear whether the reduced performance when targeting random networks, and when reprogramming to perform CIFAR-10 classification, 
was due to limitations in the expressivity of the adversarial perturbation,
or due to the optimization task in Equation \ref{eq adv obj} being more difficult in these situations. 
Disentangling limitations in expressivity and trainability will be an interesting future direction.

\subsection{Adversarial goals beyond the image domain}\label{sec beyond images}

We demonstrated adversarial reprogramming on classification tasks in the image domain. 
It is an interesting area for future research whether similar attacks might succeed for audio, video, text, or other domains and tasks. Our finding that trained networks can be reprogram to classify shuffled MNIST examples, which do not have any resemblance to images, suggest that the reprogramming across domains is likely. 

Adversarial reprogramming of recurrent neural networks (RNNs) would be particularly interesting, since RNNs (especially those with attention or memory) can be Turing complete \citep{neelakantan2015neural}. 
An attacker would thus only need to find inputs which induced the RNN to perform a number of simple operations, such as increment counter, decrement counter, and change input attention location if counter is zero \citep{minsky1961recursive}. 
If adversarial programs can be found for these simple operations, then they could be composed to reprogram the RNN to perform various tasks.


A variety of nefarious ends may be achievable if machine learning systems can be reprogrammed by a specially crafted input. 
The most direct of these is the theft of computational resources. 
For instance, an attacker might develop an adversarial program which causes the computer vision classifier in a cloud hosted photos service
 to solve image captchas and enable creation of spam accounts. 
If RNNs can be flexibly reprogrammed as 
mentioned above, this computational theft might extend to more arbitrary tasks.
A major danger beyond the computational theft is that an adversary may repurpose computational resources to perform a task which violates the code of ethics of system providers. This is particularly important as ML service providers are largely interested in protecting the ethical principles and guidelines that governs the use of their services. 

\section{Conclusion}
\label{sec conclusion}

In this work, we proposed a new class of adversarial attacks that aim to reprogram neural networks to perform novel adversarial tasks. 
Our results demonstrate for the first time the possibility of such attacks. 
They are also illustrative of 
both surprising flexibility and surprising vulnerability in deep neural networks. 
Future investigation should address the properties and limitations of adversarial programming and possible ways to defend against it.

\subsubsection*{Acknowledgments}
We are grateful to Jaehoon Lee, Sara Hooker, Simon Kornblith, Supasorn Suwajanakorn for useful comments on the manuscript. We thank Alexey Kurakin for help reviewing the code. We thank Justin Gilmer and Luke Metz for discussion surrounding the original idea.


\bibliography{adv_reprog_2018}

\begin{thebibliography}{44}
\providecommand{\natexlab}[1]{#1}
\providecommand{\url}[1]{\texttt{#1}}
\expandafter\ifx\csname urlstyle\endcsname\relax
  \providecommand{\doi}[1]{doi: #1}\else
  \providecommand{\doi}{doi: \begingroup \urlstyle{rm}\Url}\fi

\bibitem[Barabasi et~al.(2001)Barabasi, Freeh, Jeong, and
  Brockman]{barabasi2001parasitic}
Albert-Laszlo Barabasi, Vincent~W Freeh, Hawoong Jeong, and Jay~B Brockman.
\newblock Parasitic computing.
\newblock \emph{Nature}, 412\penalty0 (6850):\penalty0 894, 2001.

\bibitem[Biggio et~al.(2013)Biggio, Corona, Maiorca, Nelson, Srndic, Laskov,
  Giacinto, and Roli]{Biggio13}
Battista Biggio, Igino Corona, Davide Maiorca, Blaine Nelson, Nedim Srndic,
  Pavel Laskov, Giorgio Giacinto, and Fabio Roli.
\newblock Evasion attacks against machine learning at test time.
\newblock In \emph{Machine Learning and Knowledge Discovery in Databases -
  European Conference, {ECML} {PKDD} 2013, Prague, Czech Republic, September
  23-27, 2013, Proceedings, Part {III}}, pp.\  387--402, 2013.
\newblock \doi{10.1007/978-3-642-40994-3_25}.

\bibitem[Bratus et~al.(2011)Bratus, Locasto, Patterson, Sassaman, and
  Shubina]{bratus2011exploit}
Sergey Bratus, Michael Locasto, Meredith Patterson, Len Sassaman, and Anna
  Shubina.
\newblock Exploit programming: From buffer overflows to weird machines and
  theory of computation.
\newblock \emph{$\{$USENIX; login:$\}$}, 2011.

\bibitem[Brown et~al.(2017)Brown, Man{\'e}, Roy, Abadi, and
  Gilmer]{brown2017adversarial}
Tom~B Brown, Dandelion Man{\'e}, Aurko Roy, Mart{\'\i}n Abadi, and Justin
  Gilmer.
\newblock Adversarial patch.
\newblock \emph{arXiv preprint arXiv:1712.09665}, 2017.

\bibitem[Carlini \& Wagner(2017)Carlini and Wagner]{7958570}
N.~Carlini and D.~Wagner.
\newblock Towards evaluating the robustness of neural networks.
\newblock In \emph{2017 IEEE Symposium on Security and Privacy (SP)}, pp.\
  39--57, May 2017.
\newblock \doi{10.1109/SP.2017.49}.

\bibitem[Donahue et~al.(2014)Donahue, Jia, Vinyals, Hoffman, Zhang, Tzeng, and
  Darrell]{donahue2014decaf}
Jeff Donahue, Yangqing Jia, Oriol Vinyals, Judy Hoffman, Ning Zhang, Eric
  Tzeng, and Trevor Darrell.
\newblock Decaf: A deep convolutional activation feature for generic visual
  recognition.
\newblock In \emph{International conference on machine learning}, pp.\
  647--655, 2014.

\bibitem[Evtimov et~al.(2017)Evtimov, Eykholt, Fernandes, Kohno, Li, Prakash,
  Rahmati, and Song]{evtimov2017robust}
Ivan Evtimov, Kevin Eykholt, Earlence Fernandes, Tadayoshi Kohno, Bo~Li, Atul
  Prakash, Amir Rahmati, and Dawn Song.
\newblock Robust physical-world attacks on deep learning models.
\newblock \emph{arXiv preprint arXiv:1707.08945}, 1, 2017.

\bibitem[Ghorbani et~al.(2017)Ghorbani, Abid, and
  Zou]{ghorbani2017interpretation}
Amirata Ghorbani, Abubakar Abid, and James Zou.
\newblock Interpretation of neural networks is fragile.
\newblock \emph{arXiv preprint arXiv:1710.10547}, 2017.

\bibitem[Goodfellow et~al.(2017)Goodfellow, Papernot, Huang, Duan, Abbeel, and
  Clark]{goodfellow2017}
Ian Goodfellow, Nicolas Papernot, Sandy Huang, Yan Duan, Pieter Abbeel, and
  Jack Clark.
\newblock Attacking machine learning with adversarial examples, 2017.
\newblock URL \url{https://blog.openai.com/adversarial-example-research/}.

\bibitem[Goodfellow et~al.(2014)Goodfellow, Shlens, and
  Szegedy]{goodfellow2014explaining}
Ian~J Goodfellow, Jonathon Shlens, and Christian Szegedy.
\newblock Explaining and harnessing adversarial examples.
\newblock \emph{arXiv preprint arXiv:1412.6572}, 2014.

\bibitem[Grosse et~al.(2017)Grosse, Papernot, Manoharan, Backes, and
  McDaniel]{grosse17}
Kathrin Grosse, Nicolas Papernot, Praveen Manoharan, Michael Backes, and
  Patrick~D. McDaniel.
\newblock Adversarial examples for malware detection.
\newblock In \emph{{ESORICS} 2017}, pp.\  62--79, 2017.
\newblock \doi{10.1007/978-3-319-66399-9_4}.
\newblock URL \url{https://doi.org/10.1007/978-3-319-66399-9_4}.

\bibitem[He et~al.(2016)He, Wang, and Hopcroft]{he2016powerful}
Kun He, Yan Wang, and John Hopcroft.
\newblock A powerful generative model using random weights for the deep image
  representation.
\newblock In \emph{Advances in Neural Information Processing Systems}, pp.\
  631--639, 2016.

\bibitem[Huang et~al.(2017)Huang, Papernot, Goodfellow, Duan, and
  Abbeel]{huang2017adversarial}
Sandy Huang, Nicolas Papernot, Ian Goodfellow, Yan Duan, and Pieter Abbeel.
\newblock Adversarial attacks on neural network policies.
\newblock \emph{arXiv preprint arXiv:1702.02284}, 2017.

\bibitem[Kannan et~al.(2018)Kannan, Kurakin, and
  Goodfellow]{kannan2018adversarial}
Harini Kannan, Alexey Kurakin, and Ian Goodfellow.
\newblock Adversarial logit pairing.
\newblock \emph{arXiv preprint arXiv:1803.06373}, 2018.

\bibitem[Kolter \& Wong(2017)Kolter and Wong]{kolter2017provable}
J~Zico Kolter and Eric Wong.
\newblock Provable defenses against adversarial examples via the convex outer
  adversarial polytope.
\newblock \emph{arXiv preprint arXiv:1711.00851}, 2017.

\bibitem[Kos et~al.(2017)Kos, Fischer, and Song]{kos2017adversarial}
Jernej Kos, Ian Fischer, and Dawn Song.
\newblock Adversarial examples for generative models.
\newblock \emph{arXiv preprint arXiv:1702.06832}, 2017.

\bibitem[Krizhevsky et~al.(2012)Krizhevsky, Sutskever, and
  Hinton]{krizhevsky2012imagenet}
Alex Krizhevsky, Ilya Sutskever, and Geoffrey~E Hinton.
\newblock Imagenet classification with deep convolutional neural networks.
\newblock In \emph{Advances in neural information processing systems}, pp.\
  1097--1105, 2012.

\bibitem[{Kurakin} et~al.(2016){Kurakin}, {Goodfellow}, and
  {Bengio}]{kurakin2016mlatscale}
A.~{Kurakin}, I.~{Goodfellow}, and S.~{Bengio}.
\newblock {Adversarial Machine Learning at Scale}.
\newblock \emph{ArXiv e-prints}, November 2016.

\bibitem[Le et~al.(2011)Le, Karpenko, Ngiam, and Ng]{le2011ica}
Quoc~V Le, Alexandre Karpenko, Jiquan Ngiam, and Andrew~Y Ng.
\newblock Ica with reconstruction cost for efficient overcomplete feature
  learning.
\newblock In \emph{Advances in neural information processing systems}, pp.\
  1017--1025, 2011.

\bibitem[Lee et~al.(2009)Lee, Grosse, Ranganath, and Ng]{lee2009convolutional}
Honglak Lee, Roger Grosse, Rajesh Ranganath, and Andrew~Y Ng.
\newblock Convolutional deep belief networks for scalable unsupervised learning
  of hierarchical representations.
\newblock In \emph{Proceedings of the 26th annual international conference on
  machine learning}, pp.\  609--616. ACM, 2009.

\bibitem[Lee et~al.(2017)Lee, Bahri, Novak, Schoenholz, Pennington, and
  Sohl-Dickstein]{lee2017deep}
Jaehoon Lee, Yasaman Bahri, Roman Novak, Samuel~S Schoenholz, Jeffrey
  Pennington, and Jascha Sohl-Dickstein.
\newblock Deep neural networks as gaussian processes.
\newblock \emph{arXiv preprint arXiv:1711.00165}, 2017.

\bibitem[Li et~al.(2018)Li, Farkhoor, Liu, and Yosinski]{li2018measuring}
Chunyuan Li, Heerad Farkhoor, Rosanne Liu, and Jason Yosinski.
\newblock Measuring the intrinsic dimension of objective landscapes.
\newblock \emph{arXiv preprint arXiv:1804.08838}, 2018.

\bibitem[Lin et~al.(2017)Lin, Hong, Liao, Shih, Liu, and Sun]{lin2017tactics}
Yen-Chen Lin, Zhang-Wei Hong, Yuan-Hong Liao, Meng-Li Shih, Ming-Yu Liu, and
  Min Sun.
\newblock Tactics of adversarial attack on deep reinforcement learning agents.
\newblock \emph{arXiv preprint arXiv:1703.06748}, 2017.

\bibitem[Lin et~al.(2015)Lin, Memisevic, and Konda]{lin2015far}
Zhouhan Lin, Roland Memisevic, and Kishore Konda.
\newblock How far can we go without convolution: Improving fully-connected
  networks.
\newblock \emph{arXiv preprint arXiv:1511.02580}, 2015.

\bibitem[Liu et~al.(2016)Liu, Chen, Liu, and Song]{liu2016delving}
Yanpei Liu, Xinyun Chen, Chang Liu, and Dawn Song.
\newblock Delving into transferable adversarial examples and black-box attacks.
\newblock \emph{arXiv preprint arXiv:1611.02770}, 2016.

\bibitem[Madry et~al.(2017)Madry, Makelov, Schmidt, Tsipras, and
  Vladu]{madry2017towards}
Aleksander Madry, Aleksandar Makelov, Ludwig Schmidt, Dimitris Tsipras, and
  Adrian Vladu.
\newblock Towards deep learning models resistant to adversarial attacks.
\newblock \emph{arXiv preprint arXiv:1706.06083}, 2017.

\bibitem[Matthews et~al.(2018)Matthews, Rowland, Hron, Turner, and
  Ghahramani]{matthews2018gaussian}
Alexander G de~G Matthews, Mark Rowland, Jiri Hron, Richard~E Turner, and
  Zoubin Ghahramani.
\newblock Gaussian process behaviour in wide deep neural networks.
\newblock \emph{arXiv preprint arXiv:1804.11271}, 2018.

\bibitem[Mesnil et~al.(2011)Mesnil, Dauphin, Glorot, Rifai, Bengio, Goodfellow,
  Lavoie, Muller, Desjardins, Warde-Farley, et~al.]{mesnil2011unsupervised}
Gr{\'e}goire Mesnil, Yann Dauphin, Xavier Glorot, Salah Rifai, Yoshua Bengio,
  Ian Goodfellow, Erick Lavoie, Xavier Muller, Guillaume Desjardins, David
  Warde-Farley, et~al.
\newblock Unsupervised and transfer learning challenge: a deep learning
  approach.
\newblock In \emph{Proceedings of the 2011 International Conference on
  Unsupervised and Transfer Learning workshop-Volume 27}, pp.\  97--111. JMLR.
  org, 2011.

\bibitem[Minsky(1961)]{minsky1961recursive}
Marvin~L Minsky.
\newblock Recursive unsolvability of post's problem of" tag" and other topics
  in theory of turing machines.
\newblock \emph{Annals of Mathematics}, pp.\  437--455, 1961.

\bibitem[Moosavi-Dezfooli et~al.(2017)Moosavi-Dezfooli, Fawzi, Fawzi, and
  Frossard]{moosavi2017universal}
Seyed-Mohsen Moosavi-Dezfooli, Alhussein Fawzi, Omar Fawzi, and Pascal
  Frossard.
\newblock Universal adversarial perturbations.
\newblock In \emph{Computer Vision and Pattern Recognition (CVPR), 2017 IEEE
  Conference on}, pp.\  86--94. IEEE, 2017.

\bibitem[Neelakantan et~al.(2015)Neelakantan, Le, and
  Sutskever]{neelakantan2015neural}
Arvind Neelakantan, Quoc~V Le, and Ilya Sutskever.
\newblock Neural programmer: Inducing latent programs with gradient descent.
\newblock \emph{arXiv preprint arXiv:1511.04834}, 2015.

\bibitem[Papernot et~al.(2015)Papernot, McDaniel, Jha, Fredrikson, Celik, and
  Swami]{papernot2015limitations}
Nicolas Papernot, Patrick~D. McDaniel, Somesh Jha, Matt Fredrikson, Z.~Berkay
  Celik, and Ananthram Swami.
\newblock The limitations of deep learning in adversarial settings.
\newblock \emph{CoRR}, abs/1511.07528, 2015.

\bibitem[Papernot et~al.(2016)Papernot, McDaniel, and
  Goodfellow]{papernot2016transferability}
Nicolas Papernot, Patrick McDaniel, and Ian Goodfellow.
\newblock Transferability in machine learning: from phenomena to black-box
  attacks using adversarial samples.
\newblock \emph{arXiv preprint arXiv:1605.07277}, 2016.

\bibitem[Papernot et~al.(2017)Papernot, McDaniel, Goodfellow, Jha, Celik, and
  Swami]{papernot2017practical}
Nicolas Papernot, Patrick McDaniel, Ian Goodfellow, Somesh Jha, Z~Berkay Celik,
  and Ananthram Swami.
\newblock Practical black-box attacks against machine learning.
\newblock In \emph{Proceedings of the 2017 ACM on Asia Conference on Computer
  and Communications Security}, pp.\  506--519. ACM, 2017.

\bibitem[Peresini \& Kostic(2013)Peresini and Kostic]{peresini2013network}
Peter Peresini and Dejan Kostic.
\newblock Is the network capable of computation?
\newblock In \emph{Network Protocols (ICNP), 2013 21st IEEE International
  Conference on}, pp.\  1--6. IEEE, 2013.

\bibitem[Raghu et~al.(2016)Raghu, Poole, Kleinberg, Ganguli, and
  Sohl-Dickstein]{raghu2016expressive}
Maithra Raghu, Ben Poole, Jon Kleinberg, Surya Ganguli, and Jascha
  Sohl-Dickstein.
\newblock On the expressive power of deep neural networks.
\newblock \emph{arXiv preprint arXiv:1606.05336}, 2016.

\bibitem[Raina et~al.(2007)Raina, Battle, Lee, Packer, and Ng]{raina2007self}
Rajat Raina, Alexis Battle, Honglak Lee, Benjamin Packer, and Andrew~Y Ng.
\newblock Self-taught learning: transfer learning from unlabeled data.
\newblock In \emph{Proceedings of the 24th international conference on Machine
  learning}, pp.\  759--766. ACM, 2007.

\bibitem[Razavian et~al.(2014)Razavian, Azizpour, Sullivan, and
  Carlsson]{razavian2014cnn}
Ali~Sharif Razavian, Hossein Azizpour, Josephine Sullivan, and Stefan Carlsson.
\newblock Cnn features off-the-shelf: an astounding baseline for recognition.
\newblock In \emph{Computer Vision and Pattern Recognition Workshops (CVPRW),
  2014 IEEE Conference on}, pp.\  512--519. IEEE, 2014.

\bibitem[Shazeer et~al.(2017)Shazeer, Mirhoseini, Maziarz, Davis, Le, Hinton,
  and Dean]{shazeer2017outrageously}
Noam Shazeer, Azalia Mirhoseini, Krzysztof Maziarz, Andy Davis, Quoc Le,
  Geoffrey Hinton, and Jeff Dean.
\newblock Outrageously large neural networks: The sparsely-gated
  mixture-of-experts layer.
\newblock \emph{arXiv preprint arXiv:1701.06538}, 2017.

\bibitem[Szegedy et~al.(2013)Szegedy, Zaremba, Sutskever, Bruna, Erhan,
  Goodfellow, and Fergus]{szegedy2013intriguing}
Christian Szegedy, Wojciech Zaremba, Ilya Sutskever, Joan Bruna, Dumitru Erhan,
  Ian Goodfellow, and Rob Fergus.
\newblock Intriguing properties of neural networks.
\newblock \emph{arXiv preprint arXiv:1312.6199}, 2013.

\bibitem[TensorFlow-Slim()]{tfslim}
TensorFlow-Slim.
\newblock Tensorflow-slim image classification model library.
\newblock \url{https://github.com/tensorflow/models/tree/master/research/slim}.
\newblock Accessed: 2018-05-01.

\bibitem[{Tram{\`e}r} et~al.(2017){Tram{\`e}r}, {Kurakin}, {Papernot}, {Boneh},
  and {McDaniel}]{ensemble_training}
F.~{Tram{\`e}r}, A.~{Kurakin}, N.~{Papernot}, D.~{Boneh}, and P.~{McDaniel}.
\newblock {Ensemble Adversarial Training: Attacks and Defenses}.
\newblock \emph{ArXiv e-prints}, May 2017.

\bibitem[Ustyuzhaninov et~al.(2016)Ustyuzhaninov, Brendel, Gatys, and
  Bethge]{ustyuzhaninov2016texture}
Ivan Ustyuzhaninov, Wieland Brendel, Leon~A Gatys, and Matthias Bethge.
\newblock Texture synthesis using shallow convolutional networks with random
  filters.
\newblock \emph{arXiv preprint arXiv:1606.00021}, 2016.

\bibitem[Yosinski et~al.(2014)Yosinski, Clune, Bengio, and
  Lipson]{yosinski2014transferable}
Jason Yosinski, Jeff Clune, Yoshua Bengio, and Hod Lipson.
\newblock How transferable are features in deep neural networks?
\newblock In \emph{Advances in neural information processing systems}, pp.\
  3320--3328, 2014.

\end{thebibliography}
\bibliographystyle{iclr2019_conference}

\clearpage
\appendix
\normalsize
\onecolumn

\part*{Supplemental material}
\setcounter{figure}{0}
\renewcommand{\thefigure}{Supp. \arabic{figure}}
\setcounter{table}{0}
\renewcommand{\thetable}{Supp. \arabic{table}}

\section{Supplementary Tables}
\label{sec: supp tables}

\begin{table}[!hbt]
  \caption{Top-1 precision of models on ImageNet data}
  \label{table: ImageNet precision}
  \centering
  \begin{tabular}{ll}
    \toprule
    Model      & Accuracy \\
    \midrule
    Inception V3  &  $0.78$     \\
    Inception V4  & $0.802$     \\
    Inception Resnet V2  & $0.804$     \\
    Resnet V2 152  &  $0.778$     \\
    Resnet V2 101  &  $0.77$    \\
    Resnet V2 50 & $0.756$ \\
    Inception V3  adv. &  $0.776$     \\
    \bottomrule
  \end{tabular}
\end{table}

\begin{table}
  \caption{Hyper-parameters for adversarial program training for the square counting adversarial task. For all models, we used the Adam optimizer with its default parameters while decaying the learning rate exponentially during training. We distributed training data across a number of GPUs (each GPU receive `batch' data samples ). We then performed synchronized updates of the adversarial program parameters.}
  \label{table: hyperparams count}
  \centering
  \begin{tabular}{llllllll}
    \toprule
     ImageNet Model &  $\lambda$ & batch & GPUS & learn rate & decay & epochs/decay & steps \\
     \midrule
    Inception V3                 & $0.01$ & $50$  & $4$  &  $0.05$ & $0.96$ & $2$ & $100000$  \\
    Inception V4                & $0.01$ & $50$  & $4$  &  $0.05$ & $0.96$ & $2$ & $100000$  \\
    Inception Resnet V2    & $0.01$ & $50$  & $4$  &  $0.05$ & $0.96$ & $2$ & $100000$  \\
    Resnet V2 152            & $0.01$ & $20$  & $4$  &  $0.05$ & $0.96$ & $2$ & $100000$  \\
    Resnet V2 101           & $0.01$ & $20$  & $4$  &  $0.05$ & $0.96$ & $2$ & $60000$  \\
    Resnet V2 50           & $0.01$ & $20$  & $4$  &  $0.05$ & $0.96$ & $2$ & $100000$  \\
    \bottomrule
  \end{tabular}
\end{table}

\begin{table}
  \caption{Hyper-parameters for adversarial program training for MNIST classification adversarial task. For all models, we used the Adam optimizer with its default parameters while decaying the learning rate exponentially during training.  We distributed training data across a number of GPUs (each GPU receive `batch' data samples ). We then performed synchronized updates of the adversarial program parameters. (The Model Inception V3 adv is pretrained on ImageNet data using adversarial training method.}
  \label{table: hyperparams mnist}
  \centering
  \begin{tabular}{llllllll}
    \toprule
     ImageNet Model &  $\lambda$ & batch & GPUS & learn rate & decay & epochs/decay & steps \\
     \midrule
    Inception V3                 & $0.05$ & $100$  & $4$  &  $0.05$ & $0.96$ & $2$ & $60000$  \\
    Inception V4                & $0.05$ & $100$  & $4$  &  $0.05$ & $0.96$ & $2$ & $60000$  \\
    Inception Resnet V2    & $0.05$ & $50$  & $8$  &  $0.05$ & $0.96$ & $2$ & $60000$  \\
    Resnet V2 152            & $0.05$ & $50$  & $8$  &  $0.05$ & $0.96$ & $2$ & $60000$  \\
    Resnet V2 101           & $0.05$ & $50$  & $8$  &  $0.05$ & $0.96$ & $2$ & $60000$  \\
    Resnet V2 50           & $0.05$ & $100$  & $4$  &  $0.05$ & $0.96$ & $2$ & $60000$  \\
    Inception V3 adv.        & $0.01$ & $50$  & $6$  &  $0.05$ & $0.98$ & $4$ & $100000$  \\
    \bottomrule
  \end{tabular}
\end{table}

\begin{table}
  \caption{Hyper-parameters for adversarial program training for CIFAR-10 classification adversarial task. For all models, we used ADAM optimizer with its default parameters while decaying the learning rate exponentially during training.  We distributed training data on number of GPUS (each GPU receive `batch' data samples ). We then performed synchronized updates of the adversarial program parameters.}
  \label{table: hyperparams cifar}
  \centering
  \begin{tabular}{llllllll}
    \toprule
     ImageNet Model &  $\lambda$ & batch & GPUS & learn rate & decay & epochs/decay & steps \\
     \midrule
    Inception V3                 & $0.01$ & $50$  & $6$  &  $0.05$ & $0.99$ & $4$ & $300000$  \\
    Inception V4                & $0.01$ & $50$  & $6$  &  $0.05$ & $0.99$ & $4$ & $300000$  \\
    Inception Resnet V2    & $0.01$ & $50$  & $6$  &  $0.05$ & $0.99$ & $4$ & $300000$  \\
    Resnet V2 152            & $0.01$ & $30$  & $6$  &  $0.05$ & $0.99$ & $4$ & $300000$  \\
    Resnet V2 101           & $0.01$ & $30$  & $6$  &  $0.05$ & $0.99$ & $4$ & $300000$  \\
    Resnet V2 50           & $0.01$ & $30$  & $6$  &  $0.05$ & $0.99$ & $4$ & $300000$  \\
    \bottomrule
  \end{tabular}
\end{table}

\begin{table}
  \caption{Hyper-parameters for adversarial program training for MNIST classification adversarial task. For all models, we used the Adam optimizer with its default parameters while decaying the learning rate exponentially during training.  We distributed training data across a number of GPUs (each GPU receive `batch' data samples ). We then performed synchronized updates of the adversarial program parameters.}
  \label{table: hyperparams mnist random}
  \centering
  \begin{tabular}{llllllll}
    \toprule
     Random Model &  $\lambda$ & batch & GPUS & learn rate & decay & epochs/decay & steps \\
     \midrule
    Inception V3                 & $0.01$ & $50$  & $4$  &  $0.05$ & $0.96$ & $2$ & $100000$  \\
    Inception V4                & $0.01$ & $50$  & $4$  &  $0.05$ & $0.96$ & $2$ & $100000$  \\
    Inception Resnet V2    & $0.01$ & $50$  & $4$  &  $0.05$ & $0.96$ & $2$ & $60000$  \\
    Resnet V2 152            & $0.01$ & $20$  & $4$  &  $0.05$ & $0.96$ & $2$ & $60000$  \\
    Resnet V2 101           & $0.01$ & $20$  & $4$  &  $0.05$ & $0.96$ & $2$ & $60000$  \\
    Resnet V2 50           & $0.01$ & $50$  & $4$  &  $0.05$ & $0.96$ & $2$ & $60000$  \\
    \bottomrule
  \end{tabular}
\end{table}

\newpage


\begin{figure}
\centering
\includegraphics[width=\columnwidth]{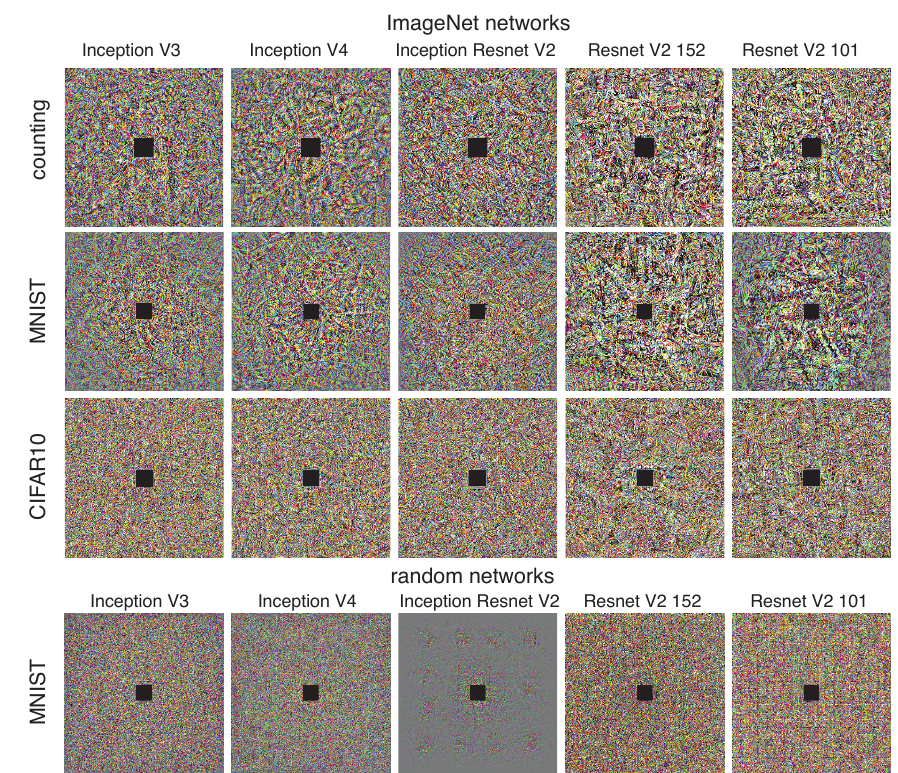}
\caption{{\bf Adversarial programs exhibit qualitative similarities and differences across both network and task.}
(a) Top: adversarial programs targeted to repurpose networks pre-trained on ImageNet to count squares in images.
Middle: adversarial programs targeted to repurpose networks pre-trained on ImageNet to function as MNIST classifiers.
Bottom: adversarial programs to cause the same networks to function as CIFAR-10 classifiers.
(b) Adversarial programs targeted to repurpose networks with randomly initialized parameters to function as MNIST classifiers.
}
\label{fig: programs}
\end{figure}

\begin{figure}
\centering
\includegraphics[width=\columnwidth]{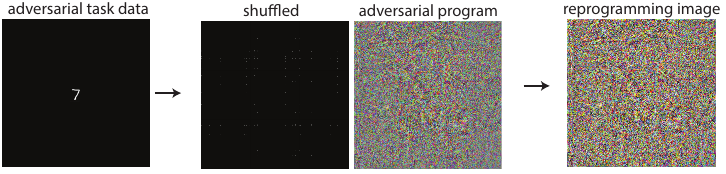}
\caption{{\bf Neural networks are susceptible to adversarial reprogramming even in cases when adversarial data and original task data are unrelated.}
The pixels in MNIST digits are shuffled. So, that the resulting image has no resemblance to any image. Then, the shuffled image is combined with the adversarial program to create a reprogramming image. This image successfully reprogram Inception V3 model to classify the shuffled digits, despite that the adversarial data (i.e., shuffled MNIST digits) being unrelated to the original data (i.e., ImageNet).
}
\label{fig: shuffled}
\end{figure}


\end{document}